\documentclass[letterpaper, 10pt]{article}

\usepackage{graphicx}
\usepackage{amsmath}
\usepackage{float}
\usepackage{amsfonts}
\usepackage[inner=0.75in, outer=0.75in, top=0.75in, bottom=1in, paperheight=11in, paperwidth=8.5in]{geometry}
\usepackage{epstopdf}
\usepackage{float}
\usepackage{amssymb}
\usepackage[utf8]{inputenc}
\usepackage[english]{babel}
\usepackage{amsthm}
\usepackage{color}
\usepackage{forest}

\usepackage{makecell}
\usepackage{booktabs}
\usepackage{cclicenses}
\usepackage{lscape,array}

\usepackage{microtype}
\usepackage{url}
\usepackage{subfiles}
\usepackage{caption}
\usepackage{subcaption}
\usepackage{parskip}
\usepackage{framed}
\usepackage[section]{placeins}

\usepackage{tikz}
\usetikzlibrary{decorations.pathmorphing}
\usetikzlibrary{decorations.markings}
\usetikzlibrary{arrows.meta,bending}
\usetikzlibrary{arrows.meta}
\usepackage{bm}

\usepackage{calrsfs}
\DeclareMathAlphabet{\pazocal}{OMS}{zplm}{m}{n}

\usepackage{adjustbox}

\usepackage{xcolor}
\usepackage{framed}

\colorlet{shadecolor}{blue!20}

\usepackage[obeyFinal,textwidth=0.55in]{todonotes}
\usepackage[hidelinks]{hyperref}
\usepackage[capitalize]{cleveref}

\theoremstyle{definition}

\begin{document}	

\setcounter{page}{1}
\pagenumbering{gobble}

\title{\bf On the Detection and Quantification of Nonlinearity via Statistics of the Gradients of a Black-Box Model}	
\author{G.\ Tsialiamanis\textsuperscript{1}, C.R.\ Farrar\textsuperscript{2}\\
        \textsuperscript{1}Dynamics Research Group, Department of Mechanical Engineering, University of Sheffield \\
        Mappin Street, Sheffield S1 3JD, UK\\
        \textsuperscript{2}Engineering Institute, MS T-001, Los Alamos National Laboratory, Los Alamos, NM 87545, USA
	   }
	\date{}
    \maketitle

\section*{Abstract}
Detection and identification of nonlinearity is a task of high importance for structural dynamics. On the one hand, identifying nonlinearity in a structure would allow one to build more accurate models of the structure. On the other hand, detecting nonlinearity in a structure, which has been designed to operate in its linear region, might indicate the existence of damage within the structure. Common damage cases which cause nonlinear behaviour are breathing cracks and points where some material may have reached its plastic region. Therefore, it is important, even for safety reasons, to detect when a structure exhibits nonlinear behaviour. In the current work, a method to detect nonlinearity is proposed, based on the distribution of the gradients of a data-driven model, which is fitted on data acquired from the structure of interest. The data-driven model selected for the current application is a neural network. The selection of such a type of model was done in order to not allow the user to decide how linear or nonlinear the model shall be, but to let the training algorithm of the neural network shape the level of nonlinearity according to the training data. The neural network is trained to predict the accelerations of the structure for a time-instant using as inputs accelerations of previous time-instants, i.e. one-step-ahead predictions. Afterwards, the gradients of the output of the neural network with respect to its inputs are calculated. Given that the structure is linear, the distribution of the aforementioned gradients should be unimodal and quite peaked, while in the case of a structure with nonlinearities, the distribution of the gradients shall be more spread and, potentially, multimodal. To test the above assumption, data from an experimental structure are considered. The structure is tested under different scenarios, some of which are linear and some of which are nonlinear. More specifically, the nonlinearity is introduced as a column-bumper nonlinearity, aimed at simulating the effects of a breathing crack, and at different levels, i.e. different values of the initial gap between the bumper and the column. Following the proposed method, the statistics of the distributions of the gradients for the different scenarios can indeed be used to identify cases where nonlinearity is present. Moreover, via the proposed method one is able to quantify the nonlinearity by observing higher values of standard deviation of the distribution of the gradients for lower values of the initial column-bumper gap, i.e. for ``more nonlinear'' scenarios.

\textbf{Key words: Structural health monitoring (SHM), structural dynamics, nonlinear dynamics, machine learning, neural networks.}

\section*{Introduction}
\label{sec:intro}

In the pursuit of making everyday life safer, humans have extensively tried to model the environment around them. Structures are an important part of the environment, in which humans live. They are man-made and should be safe throughout their lifetime. Structures are exposed to numerous environmental factors, which may cause them to fail. Moreover, during operation, structures are subjected to dynamic loads, which, in time, may cause failure. Such failures will most probably result in economic damage to society and may even result in loss of human lives. Therefore, for the purpose of maintaining structures safe, the field of \textit{structural health monitoring} (SHM) \cite{Farrar} has emerged.

\par

The discipline of SHM is the subdiscipline of structural dynamics, which focuses on using data acquired from sensors to evaluate the condition of a structure. There are several tasks that are performed during an SHM project. A convenient categorization of these tasks has been proposed by Rytter in \cite{rytter1993vibrational}, extended in \cite{Farrar} and is given by the hierarchical structure,

\begin{enumerate}
	\item Is there damage in the system ({\em existence})?
	\item Where is the damage in the system ({\em location})?
	\item What kind of damage is present ({\em type/classification})?
	\item How severe is the damage ({\em extent/severity})?
	\item How much useful (safe) life remains ({\em prognosis})?
\end{enumerate}

\par

The order of the tasks in the above hierarchy may also be viewed as being defined according to their difficulty. The first step, that of identifying whether damage exists in a structure, is considered the most simple one. Several approaches exist, but a quite common one is \textit{outlier detection} \cite{Barnett}. Such approaches could range from fitting simple probability density functions to the data to fitting an \textit{autoencoder neural network} (ANN) to the data to calculate a novelty index \cite{worden2003experimental}. A general strategy for this task is to define a baseline model of the structure, which accurately explains the healthy state, and indicate any state of the structure, which may not be adequately explained by this model, as a potentially damaged state \cite{sohn2001damage}. 

\par

The second and the third steps of the hierarchy, the localization or the classification of damage, require more specialized tools than the case of detecting damage. As in many cases of modelling systems, there are two types of models to adopt to localize damage, \textit{physics-based} models (e.g.\ \textit{finite element models} \cite{bathe2006finite}) and \textit{data-driven} or \textit{machine learning} models \cite{Bishop2}. The first category refers to models which are built according to one's understanding of the physics of damage, e.g. cracks \cite{agathos2018multiple}. The second type of models, machine learning models, are built using mainly data and algorithms which perform pattern recognition to indicate the position of the damage within the structure \cite{manson2003experimental}. 

\par

The next two steps are considered to be the most difficult ones. To define the extent of damage or how much useful life the structure has, one should have adequate knowledge about the physics of the mechanism of the specific type of damage that affects the structure \cite{corbetta2018optimization}. In order to perform these processes in a data-driven manner, one should have data available of the evolution of damage of the structure, which might be difficult to acquire. Therefore, a way to deal with problems of the fourth and fifth step of Rytter's hierarchy might be to follow a \textit{population-based structural health monitoring} (PBSHM) framework \cite{gardner2022population}, in order to exploit data from structures in a population, which have already failed, to make predictions about the evolution of damage of existing structures in the same population. In every case, these steps (especially the last one) involve many uncertain parameters, such as the environmental and operational conditions of the structure in the future, making them quite complicated tasks.

\par

All of the steps of the hierarchy have been approached in both a physics-based and a data-driven way and each of these ways has its advantages and disadvantages. On the one hand, physics-based models are based on the underlying physics of the structure and the damage mechanism, therefore, if the formulation of the physics matches or is ``close'' to the real underlying physics, the models should make accurate predictions about it. However, because of the uncertainties of the environment, capturing accurately the underlying physics for a structure does not ensure that other structures - even identical ones - shall behave the same way. 

\par

Data-driven models, on the other hand, do not require definition of analytical expressions regarding the underlying physics of a structure. They rely exclusively on data. An apparent drawback of such models is that it might be challenging to consider them in cases where data from a structure are not available. Moreover, since the main object of study of SHM is damage, data-driven methods are even more difficult to be applied, because, when structures are damaged, they are often quickly repaired, making the access to data from damaged structures even more restricted. Even if one acquires data from the damaged structure, the models trained according to these data may not be efficient for the repaired structure, because the structural parameters may change, as a result of the repair procedure \cite{gardner2022challenges}.

\par

The current work is mainly focused on the first step of the hierarchy, that of identifying whether a structure is damaged or not. The approach followed herein is motivated by \cite{farrar2010complexity} and \cite{worden2019applying}, where it is assumed that a damaged structure should provide more complex data than a healthy structure - the complexity being measured by several complexity metrics. The approach proposed herein is that of a metric of complexity of the gradients of a neural network model that sufficiently explains the data acquired from a structure. As it will be discussed, the proposed metric is based on the statistics of the gradients of the model and provides a way of discriminating states of the structure that indicate nonlinear behaviour. Moreover, the proposed metric can be calculated for different degrees of freedom of the structure so it may also be used as a tool for the second step of the hierarchy, the localization of damage. Finally, being a scalar metric, in the example presented here, it's magnitude may also be used as an indication of ``how intensely nonlinear'' a structure is.

\par

The layout of the paper is the following. In the second section, a brief discussion is provided about machine learning in structural dynamics and more specifically SHM and the reasons that it has been widely exploited in many disciplines, including engineering. In the third section, the dataset which is used in the current work is presented and aspects of the specific dataset are discussed. In the fourth section, the proposed methodology is presented and the results of applying it on the dataset. In the final section, conclusions are drawn regarding the proposed algorithm and future steps are discussed.

\section*{Machine learning for structural dynamics}

In recent years, machine learning has been widely developed for computer science and many other disciplines as well. One would argue that machine learning gets so much attention because of the impressiveness of the results of its algorithms. For example, machine learning algorithms have achieved generation of real-looking images \cite{goodfellow2014generative}, translation of text into artificially generated images \cite{ramesh2022hierarchical} and prediction of the structure of proteins \cite{jumper2021highly}, achievements that before the rise of machine learning would have been considered very difficult or even impossible. However, an equally-important reason of the success of machine learning is that it has provided an alternative solution to problems that classical approaches have failed. 

\par

One of the problems that machine learning has solved quite successfully is that of image recognition \cite{lecun1995convolutional}. A basic application of image recognition is to simply classify images in classes, regarding the content of the image. The advantage of using machine learning is that the model which takes an image as an input and predicts the class of the image is learnt exclusively from the available data. A classical approach to this problem may have been very difficult to be attained. One may have had to manually identify patterns in the images that point towards each class. Afterwards, one would have to define, also manually, a model that quantifies how plausible it is for every image to belong to a class, according to these patterns. Machine learning and specifically the \textit{convolutional neural network} (CNN) \cite{lecun1995convolutional} provided a solution to these problem without any need of human intervention in the parameters of the model (only the \textit{hyperparameters} of a machine learning algorithm need manual tuning).

\par

Similarly, for SHM, one of the problems of Rytter's hierarchy can be solved via the use of classification machine learning algorithms. An excellent example of such a solution is presented in \cite{manson2003experimental}, where a neural network is built to predict the category of damage on a wing of an aircraft. The damage was simulated by removing panels from the wing and data were acquired during nine different damage cases. The model is trained according to data corresponding to the damaged states and it achieves quite high accuracy in classifying unseen data into the nine damage classes. The image recognition and the damage classification applications reveal a major advantage of machine learning, the bypassing of manually defining a model to perform the desired task. Especially for the damage classification task, one would need extended understanding of the way that the removed panels should affect the recorded data, something that might be quite structure-specific as results in \cite{gardner2022challenges} reveal. However, because of the uncertainties of such a phenomenon, machine learning proves a far more convenient and effective solution.

\par

The two examples mentioned in the previous paragraphs refer to problems whose solution in a traditional way may have been infeasible, i.e.\ the definition of closed form equations that would classify the images into the desired classes or the sensor signals into the damage classes. However, machine learning can provide solutions in a similar manner in case of problems for which one may have available equations. An example of such a problem is that of system identification \cite{rogers2020application}. Quite often, one may have a parametric set of equations for a system but the set of equations may suffer from \textit{epistemic uncertainty}, i.e.\ the equations may not resemble the completely-true underlying physics of the problem. Therefore, such a set of equations may not be able to sufficiently explain the phenomenon and make accurate predictions. In such cases, a \textit{black-box} modelling approach can be followed. A black-box model is simply a model that learns exclusively from data the relationship between some input and some output quantities of interest. A characteristic example of such a model is a neural network \cite{Bishop2}.

\par

In the current work, a functionality of neural networks which has become popular in the community of scientific machine learning is exploited. This functionality refers to the calculation of gradients of neural network models with respect to their inputs. Differentiation is a very important part of defining and training a neural network, because \textit{backpropagation}, the algorithm used to train a neural network, is based on calculating the derivatives of a \textit{loss function} with respect to the tunable parameters of the network. Recently, the calculation of derivatives has been extended to the inputs of the neural network because of the definition of \textit{physics-informed neural networks} (PINN) \cite{raissi2019physics}. For such a type of neural networks, the researchers impose their knowledge of the underlying physics of the studied system as part of the loss function. This knowledge often involves relationships of the derivatives of the defined model, therefore, the calculation of derivatives with respect to input quantities has been utilized. In the case of PINNs the gradient is used as a means of imposing physical knowledge into the model. However, in the current work the gradients are studied after training the model as a way of analyzing the model, aiming at a better study and higher explainability of the model.

\section*{A three-storey building dataset}

\begin{figure}[H]
    \centering
    \includegraphics[scale=0.7]{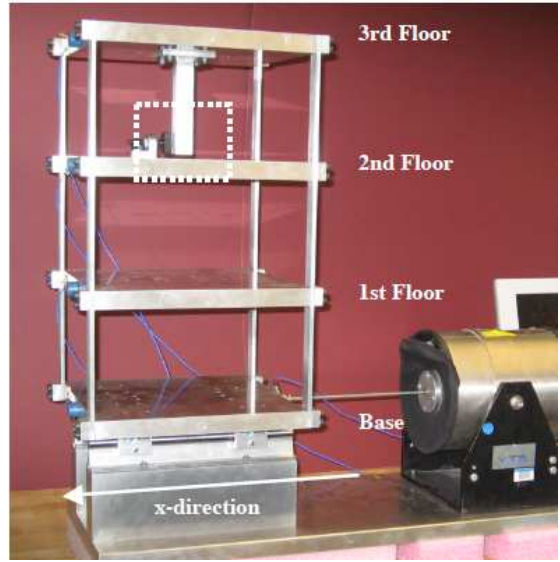}
    \caption{Experimental set-up of three-floor and the bumper nonlinearity between the second and third floor shown in the dashed box \cite{figueiredo2009structural}.}
    \label{fig:LANL_bookshelf}
\end{figure}

The dataset which is used in the current work comes from a three-storey structure, shown in Figure \ref{fig:LANL_bookshelf}, which was tested in the laboratory. The structure was considered in 17 different states, but the 14 of those are of interest of the current work. State $1$ is considered the baseline state, states $2-9$ are considered to be undamaged and states $10-14$ are considered the damaged ones. The difference between states $2-9$ from the baseline state is regarding the stiffness of different columns of the structure or some added mass on a floor. For example, state $3$ has an extra $1.2$ kg on the first floor and state 6 has a stiffness reduction on the front column of the first floor. In large scale structures, such stiffness reductions could be observed in cases of temperature differences between two parts of the structure, e.g.\ a bridge whose deck is heated by the sun, but the underneath structure is heated slower. The damage for states $10-14$ is simulated by engaging the bumper and the column between the second and the third floor, shown within the dashed box in Figure \ref{fig:LANL_bookshelf}. A damage type, which could be simulated by such a column-bumper setup could be a breathing crack. The difference between the damaged states is the width of the initial gap between the column and the bumper and for higher number of state, the gap becomes smaller, which can be considered an increase in how harsh the nonlinearity of the structure is, because the smaller the gap, the more often the two elements will collide and probably with higher velocity. Namely, the gap was $0.20$mm, $0.15$mm, $0.13$mm, $0.10$mm and $0.05$mm for states $10-14$ respectively.

The structure was excited using a white noise signal on its base. The acquired data came from accelerometers placed on each floor and the base. For every state of the structure, the experiment was repeated $50$ times and for every repetition the excitation and the recording of the data lasted $25.6$ seconds and the data were recorded in a sampling frequency of $320$ Hz, resulting in $8192$ data points for each one of the $50$ repetitions. Therefore, for every state, a dataset of $409600$ points is available.

\par

The aim of the current work is to define a metric which shall be used to identify the existence of damage and which shall have a higher value for greater extend of damage. As mentioned, in the current dataset, the damaged states are considered the ones which have a nonlinearity. Furthermore, the nonlinearity of the structure is considered to be increasing as the gap between the column and the bumper is decreasing. On the contrary, states $2-9$ are different compared to the baseline state $1$, but are still linear and not considered as damaged. As a result, the desired metric should be a metric of nonlinearity which should have higher value as the structure becomes more nonlinear or, in the current case, as the gap between the bumper and the column becomes smaller. As a first step, a model is desired, which can be used to make accurate predictions and be studied in order to define a metric with the aforementioned properties.

\par

Following a physics-based framework, such a model, which would be fitted to the data, would have a predefined form, for example,
\begin{equation}
    [M]\{\ddot{\mathbf{y}}\} + [C]\{\dot{\mathbf{y}}\} + [K]\{\mathbf{y}\} = {F}
\end{equation}
where $[M]$, $[C]$, $[K]$ are the mass, damping and stiffness matrices respectively, $\{ \ddot{\mathbf{y}}\}$, $\{\dot{\mathbf{y}}\}$, $\{\mathbf{y}\}$ are the acceleration, velocity and displacement vectors and ${F}$ is the forcing vector of the system. A major problem of such an approach is that if the predefined equation does not match the underlying physics of the structure, the model may not be sufficient to make accurate predictions. Although the structure is considered linear for states $1-9$, there might be nonlinearities for example in the joints or the damping.

\par

Another approach for a model that can be studied in order define the desired metric, is to define a machine learning model. For this purpose, in the current work, a neural network is chosen as such a model. The reasons to use a neural network are two. First, a neural network is a universal approximator \cite{csaji2001approximation}, making it a great model in the case of unknown underlying physics. An example of a model which does not have such a property is a polynomial regression model of predefined order. The second reason is that a neural network can be recalibrated from its baseline state according to data from a new state. The desired approach herein is to define a model for the baseline state and to recalibrate it for new data. Following this approach, the change of the model can be studied and compared to the baseline model to define the aforementioned metric.

\section*{Statistics of the model gradients as a nonlinearity metric}

For a single-degree-of-freedom system, a general equation that can be considered for dynamic systems is given by,
\begin{equation}
    \label{eq:general_dynamic}
    m \ddot{y} + c \dot{y} + k y + g(y, \dot{y}) = F(t)
\end{equation}
where $m$, $c$, $k$ are the mass, damping and stiffness of the oscillator, $\ddot{y}$, $\dot{y}$, $y$ are the acceleration, velocity and displacement of the system, $F(t)$ is the force signal applied to the system and $g(y, \dot{y})$ is a function of nonlinear terms of $y$. A common way to solve such an equation is in a discrete-time framework, by replacing $\ddot{y}$ and $\dot{y}$ by their finite difference approximation, i.e. $\dot{y}=\frac{y_{t} - y_{t-1}}{dt}$ and $\ddot{y}=\frac{\dot{y}_{t} - \dot{y}_{t-1}}{dt}$. The solution then is a \textit{one-step-ahead} (OSA) model has the form,
\begin{equation}
    \label{eq:discrete_solution}
    y_{t} = f(y_{t-1}, y_{t-2},... y_{t-l})
\end{equation}
where $f$ is a model and $l$ is the \textit{lag}, which is number of timesteps before the timestep $t$ that are used as inputs to the model. For the linear case, the equation above becomes $y_{t}=(2 - \frac{c\Delta t}{m} - \frac{k\Delta t^2}{m})y_{t-1} + (\frac{c\Delta t}{m} - 1)y_{t-2} + \frac{\Delta t^2}{m} F_{t-1}$ \cite{Farrar}.

\par

From equation (\ref{eq:discrete_solution}) the gradients of $y_{t}$ with respect to $y_{t-i}$ can be calculated as,
\begin{equation}
    \frac{\partial y_{t}}{\partial y_{t-i}}|_{y_{0}} = \frac{\partial f}{\partial y_{t-i}}|_{y_{0}} \quad i=1, 2,... l
\end{equation}
where $y_{0}$ is the value of $y_{t-i}$ for which the derivative is calculated. The distribution of these derivatives is a meaningful object which characterizes the system. For the linear case, the value of $g(y, \dot{y})$ of equation (\ref{eq:general_dynamic}) is zero, making the relationship of equation ($\ref{eq:discrete_solution}$) linear and the value of the aforementioned gradients constant. The distribution of the gradients should be a Kronecker delta. However, since the model $f$ is a statistical model and noise will be present in the data, the distribution is expected to be a quite narrow Gaussian-like distribution.

\par

For a wide range of types of nonlinearity, the distribution of the gradients is expected to spread. Quite often, nonlinear systems exhibit nonlinear behaviour for larger values of displacements, e.g.\ the Duffing oscillator. In these cases, the value of $g(y, \dot{y})$ of equation (\ref{eq:general_dynamic}) is not equal to zero, but as the magnitude of $y$ increases, the contribution of $g$ to the movement of the system becomes more evident. For the example of the Duffing oscillator $g(y, \dot{y}) = k_{3}y^3$, therefore the derivative will have constant terms, because of the linear part of the system, and nonlinear terms proportional to $y^{2}$, which for small values of $y$ is almost equal to zero, but for larger values affects the value of the derivative. As a result, the distribution of the derivatives will not be as narrow as in the linear case.

\par

Using the available dataset from the three-storey building, an OSA model can be defined using a neural network. The model can be defined as,
\begin{equation}
    \label{eq:model}
    \{\ddot{\mathbf{y}}_{t}\} = f(\{\ddot{\mathbf{y}}_{t-1}\}, \{\ddot{\mathbf{y}}_{t-2}\},...\ \{\ddot{\mathbf{y}}_{t-l}\})
\end{equation}
where $\{\ddot{\mathbf{y}}_{t}\} = [y^{1}_{t}, y^{2}_{t}, y^{3}_{t}, y^{4}_{t}]$ is the vector of accelerations of the four degrees of freedom for timestep $t$, $f$ is the neural network model and $l$ is the \textit{lag}, which is number of timesteps before the timestep $t$ that are used as inputs to the model. To emulate a real situation where the forcing vector is not available to use for modelling, it is not used as an input to the model, however the accelerations of the base can be considered the forcing of the rest of the building, as in an earthquake situation.

\par

Having defined a model such as the one defined in equation (\ref{eq:model}), allows studying its gradients with respect to its inputs. More specifically, for the acceleration of the $i$th degree of freedom, the derivative with respect to the acceleration of the same degree of freedom but of the previous timestep is defined as,
\begin{equation}
    \frac{\partial y^{i}_{t}}{\partial y^{i}_{t-1}}|_{y^{i}_{0}} = \frac{\partial f^{i}}{\partial y^{i}_{t-1}}|_{y^{i}_{0}}
\end{equation}
where $f^{i}$ is the $i$th output of the neural network model of equation (\ref{eq:model}). The quantity of the above equation is calculated for a specific value of $y^{i}_{t-1} = y^{i}_{0}$. Therefore, by calculating the derivative for several values of $y^{i}_{t-1}$ one can calculate the distribution of these derivatives.

\par

For the three-storey structure, a simple feedforward neural network is chosen as the model, which will be fitted on the available data. The considered neural network had three layers, an input layer, a hidden layer and an output layer. The dimensionality of the input layer is defined by the size of the lag, which is one of the hyperparameters of the algorithm and should be optimized. The size of the output layer can be equal to the number of degrees of freedom of the system, according to equation (\ref{eq:model}), but to allow the model to be specialized separately to each degree of freedom, in the current work, different models are trained for the accelerations of each degree of freedom, i.e.\ $\ddot{y}^{i}_{t} = f(\{\ddot{\mathbf{y}}_{t-1}\}, \{\ddot{\mathbf{y}}_{t-2}\},...\ \{\ddot{\mathbf{y}}_{t-l}\})$.

\par

The size of the hidden layer is also a hyperparameter. A common approach for the size of the hidden layer is to train the neural network using a part of the available dataset, called the training dataset, and calculate the error on a second part of the dataset, called the validation dataset. Then the optimal size of the hidden layer is considered one with which the minimum error on the validation dataset is achieved. Although this is an effective strategy to optimize the size of the hidden layer, in the current work, it is predefined by the authors. The size is predefined because the model is fitted on the baseline state data and then shall be recalibrated according to the data from the new states. Having the same model offers the chance for a comparison between the distribution of the gradients of the baseline state and that of the new states. Therefore, the size of the hidden layer should be chosen in a way to allow the neural network to have approximation capabilities for a wide range of linear and nonlinear states. The size is chosen to be $100$ neurons, which is considered a descent size. In every case, the model is tested on unseen data to ensure that the error on these data is not high, i.e. it is not overfitted to the training data.

\par

As mentioned, proper training requires splitting the complete dataset in a training and a validation dataset. In order to test the algorithm on unseen data, a third part is considered, the testing dataset. The whole procedure followed to train the baseline model was to split the whole dataset of state $1$ into the three aforementioned parts. Afterwards different values for the lag were studied, to train the model on the training dataset and to pick the one model that performs best on the validation dataset. It was noted that a lag value of $2$ was sufficient for the model to perform satisfactorily on the validation dataset. 

\par

The performance of the model was examined using the \textit{normalized mean-square error} (NMSE), given by $\frac{100}{N\sigma_{y}^{2}}\sum_{i=1}^{N}(\hat{y}_{i} - y_{i})^{2}$, where $N$ is the number of available samples in the dataset, $\sigma_{y}$ is the standard deviation of the target values, $\hat{y}_{i}$ are the predictions of the model, and $y_{i}$ are the actual target values. The NMSE is a convenient measure of error in regression problems, since it provides an objective measure of the accuracy, regardless of the scale of the data. NMSE values close to $100\%$ indicate that the model does no better than simply using the mean value of the data, while the lower the value the better the model is calibrated. From experience, values of NMSE lower than $5\%$ indicate a well-fitted model, and values lower than $1\%$ show an excellent model. For the applications presented here, the NMSE was for every state lower than $5\%$ for all three datasets (training, validation testing), indicating that the models were not overfitted.

\par

Another aspect of training, which is considered important in the current work was the use of \textit{regularization} \cite{krogh1991simple}. Regularization is often used to prevent overfitting by enforcing smoother mappings between the input and the output quantities of the neural network. In the current work, its benefit is considered to be the smoothness of the mapping, which shall allow the gradients to smoothly change and provide more smooth distributions. For the same reason, the activation functions of the neural network are \textit{hyperbolic tangent} (tanh) activation functions for the hidden layer and a linear activation function for the output layer.

\par

After fitting the model to the baseline state, for every new state, the model was recalibrated. This procedure is simply the training of the neural network considering as the initial state of its trainable parameters the trained-on-the-baseline-data model. After retraining for each different state, the gradients were calculated for the accelerations of every degree of freedom with respect to every input variable (the lagged accelerations). For these distributions three moments were calculated. The standard deviation, the skewness and the the kurtosis. The standard deviation and the inverse of the kurtosis exhibited good results and are presented herein, the skewness did not exhibit any worth-presenting results. More specifically, the mean values of the moments are used, given by,
\begin{equation}
    \label{eq:metrics}
    \bar{M}^{i} = \frac{1}{D} \sum_{j=1}^{D}{M}[\frac{\partial f^{i}}{\partial y^{j}_{t-1}}|_{y^{j}_{0}}] \quad i=1, 2, ... n_{dof}
\end{equation}
where $M$ is a moment (standard deviation or inverse kurtosis herein), $D$ is the dimensionality of the input, $n_{dof}$ is the number of degrees of freedom of the system and $y^{j}_{0}$ are the points for which the gradients are calculated - these points in the current work are all the available points in the three-storey-building dataset. Essentially, the moments are calculated separately for the distributions of the gradients with respect to every input quantity and the average of these values is used as a metric.

\par

\begin{figure}[H]
\centering
\begin{subfigure}{.5\textwidth}
  \centering
  \includegraphics[scale=0.62]{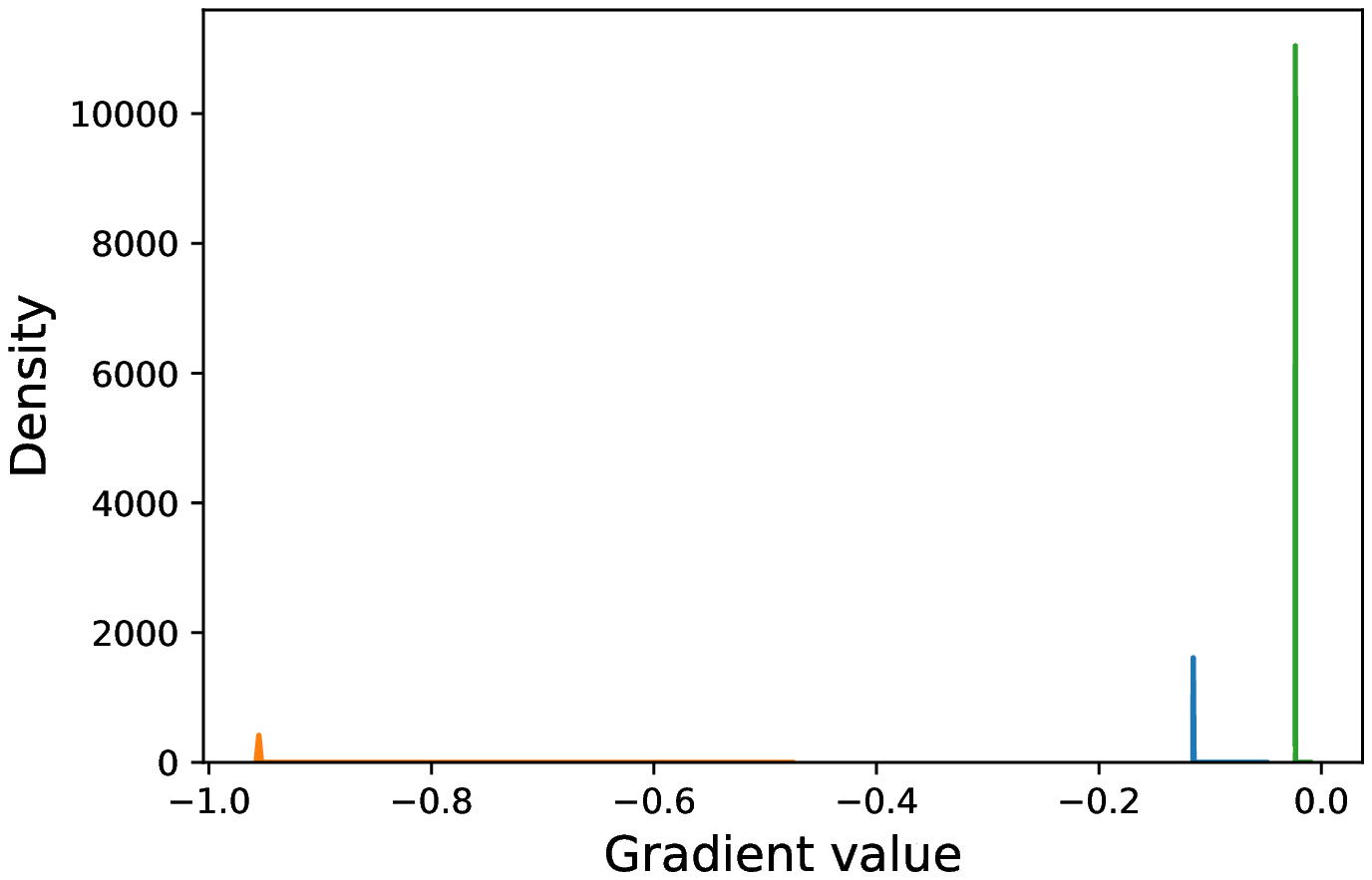}
  \label{fig:baseline_example}
\end{subfigure}%
\begin{subfigure}{.5\textwidth}
  \centering
  \includegraphics[width=\linewidth]{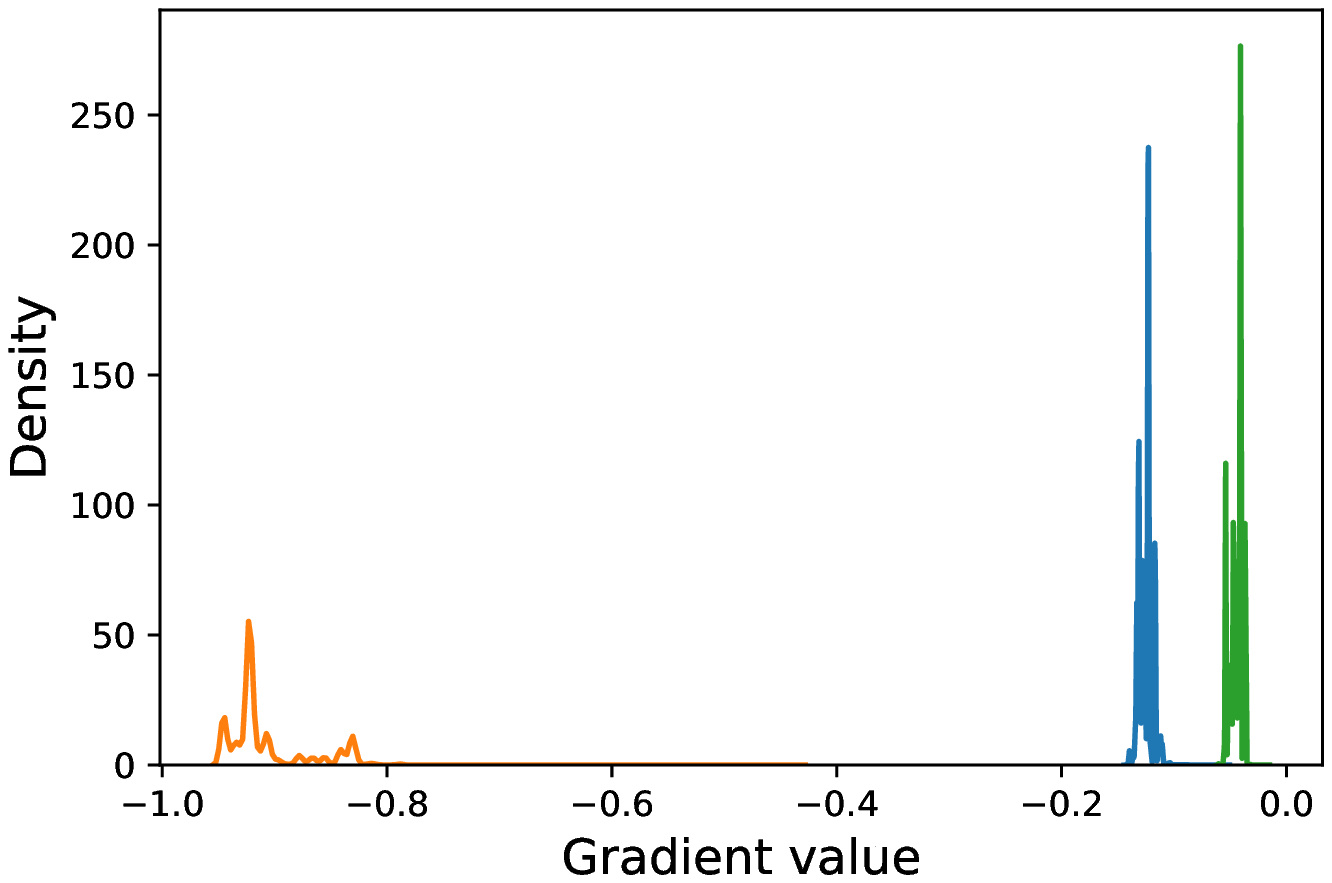}
  \label{fig:state_14_example}
\end{subfigure}
\caption{Distribution of the gradients of the model for the predictions of the third floor with respect to the accelerations of the previous timestep of the base (blue line), the first floor (orange line) and the second floor (green line). The distributions are shown for two cases, for the baseline state (left) and for state $14$ (right), the most nonlinear case.}
\label{fig:grads_example}
\end{figure}

At first, the aforementioned distributions are presented to provide a visual confirmation of the intuition presented in the previous paragraphs. Since samples of the gradients are available, a probability density function (PDF) needs to be defined for the purpose of visualizing the distributions. The PDF was calculated using a kernel density estimation and the Silverman's algorithm \cite{silverman1981using}. The results for the baseline state (state $1$) and for the most nonlinear case (state $14$) are presented in Figure \ref{fig:grads_example}. From the plots, it is clear that for the undamaged state the distribution of the gradients is much more peaked than in the case of the nonlinear structure, where the expected spread of the distributions is clearly observed.

\begin{figure}[H]
\centering
\begin{subfigure}{.5\textwidth}
  \centering
  \includegraphics[scale=0.62]{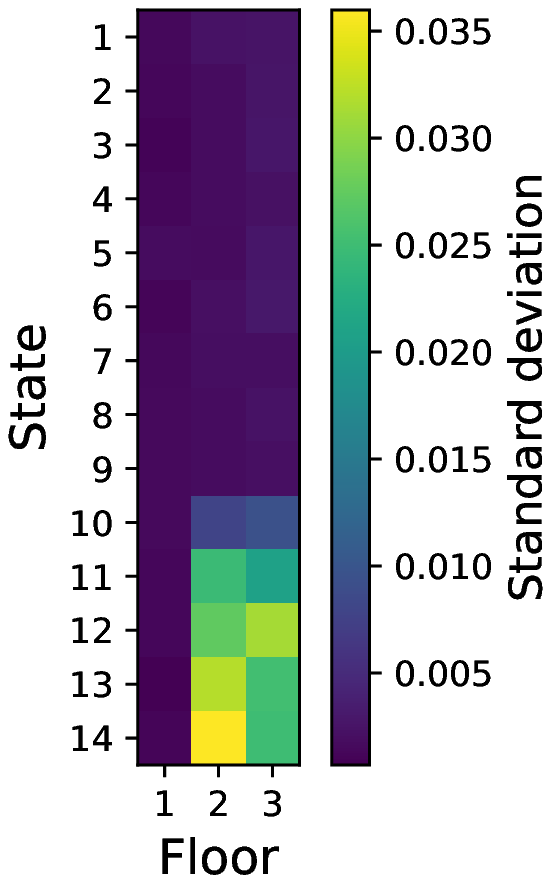}
  \label{fig:stds}
\end{subfigure}%
\begin{subfigure}{.5\textwidth}
  \centering
  \includegraphics[width=\linewidth]{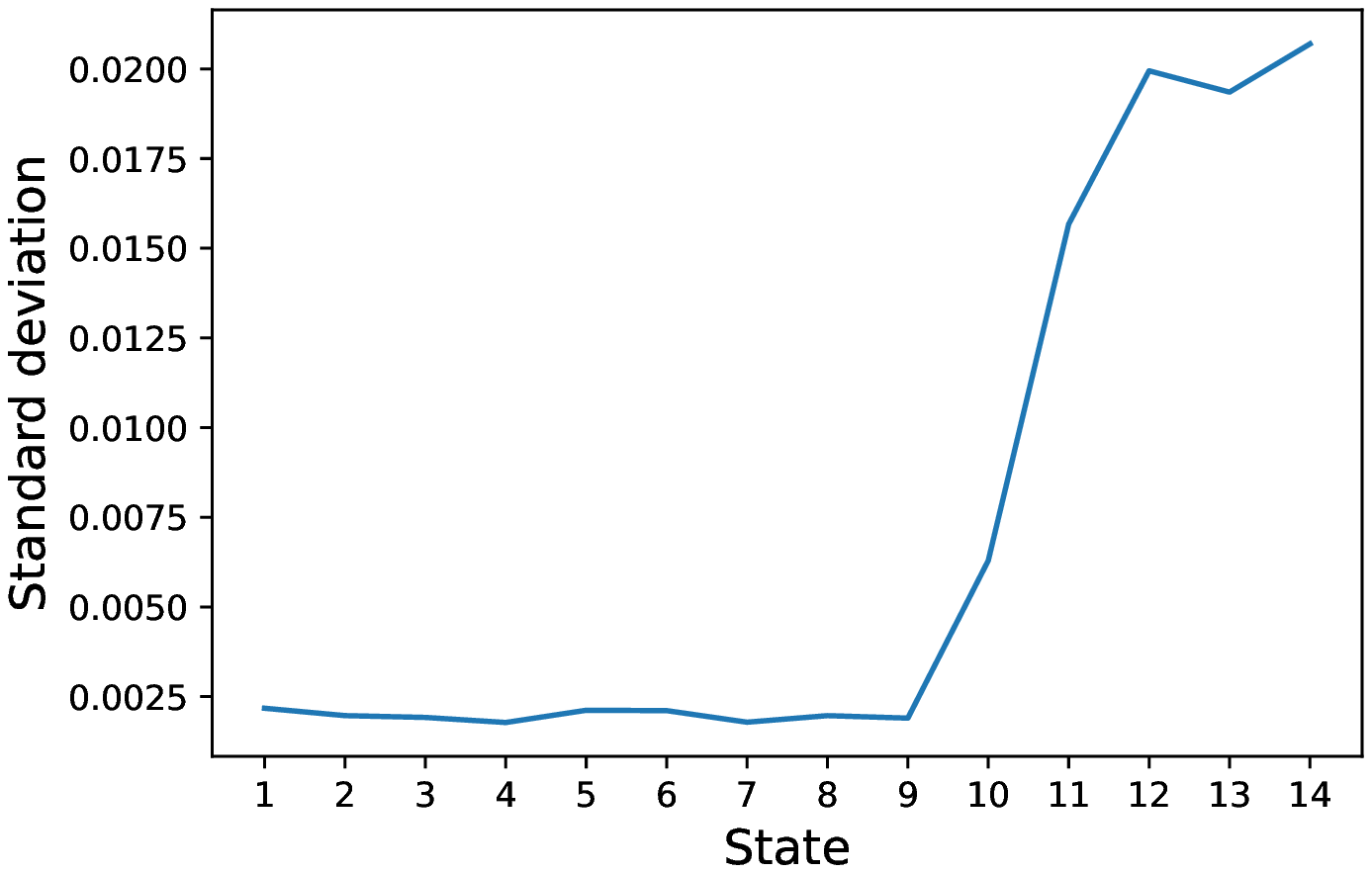}
  \label{fig:mean_stds}
\end{subfigure}
\caption{The values of the metric of equation ($\ref{eq:metrics}$) using the standard deviation as $M$, for the accelerations of the three floors of the structure (left) and the average value of all three floors of the metric for the different states (right).}
\label{fig:stds_mean_stds}
\end{figure}

Subsequently, using equation (\ref{eq:metrics}) two metrics  are calculated. The first metric is the standard deviation of the distributions of the gradients. The standard deviation is a metric of spreadness of the distribution. Higher values of standard deviation mean a more spread distribution. In Figure \ref{fig:stds_mean_stds} on the left, using equation (\ref{eq:metrics}), the average standard deviation of the the model of the acceleration of the three floors is shown for different states. On the right of the figure, the mean value of the rows of the left-side plot are shown. The results reveal that using the standard deviation as a metric, the nonlinear states can be identified. Moreover, a monotonic increase on the value of the metric is observed, as the column-bumper gap becomes smaller, except for state $13$, where a small decrease is observed, compared to states $12$ and $14$. Another interesting aspect of the metric is that its value is increasing mainly for the second and third floor, where the nonlinearity is introduced, making it a candidate metric for localization of damage.

\begin{figure}[H]
\centering
\begin{subfigure}{.5\textwidth}
  \centering
  \includegraphics[scale=0.62]{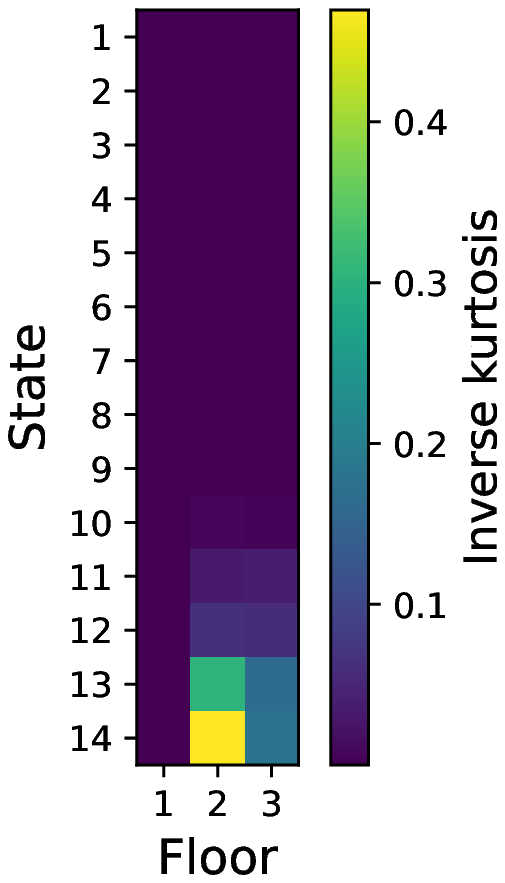}
  \caption{}
  \label{fig:inv_kurts}
\end{subfigure}%
\begin{subfigure}{.5\textwidth}
  \centering
  \includegraphics[width=\linewidth]{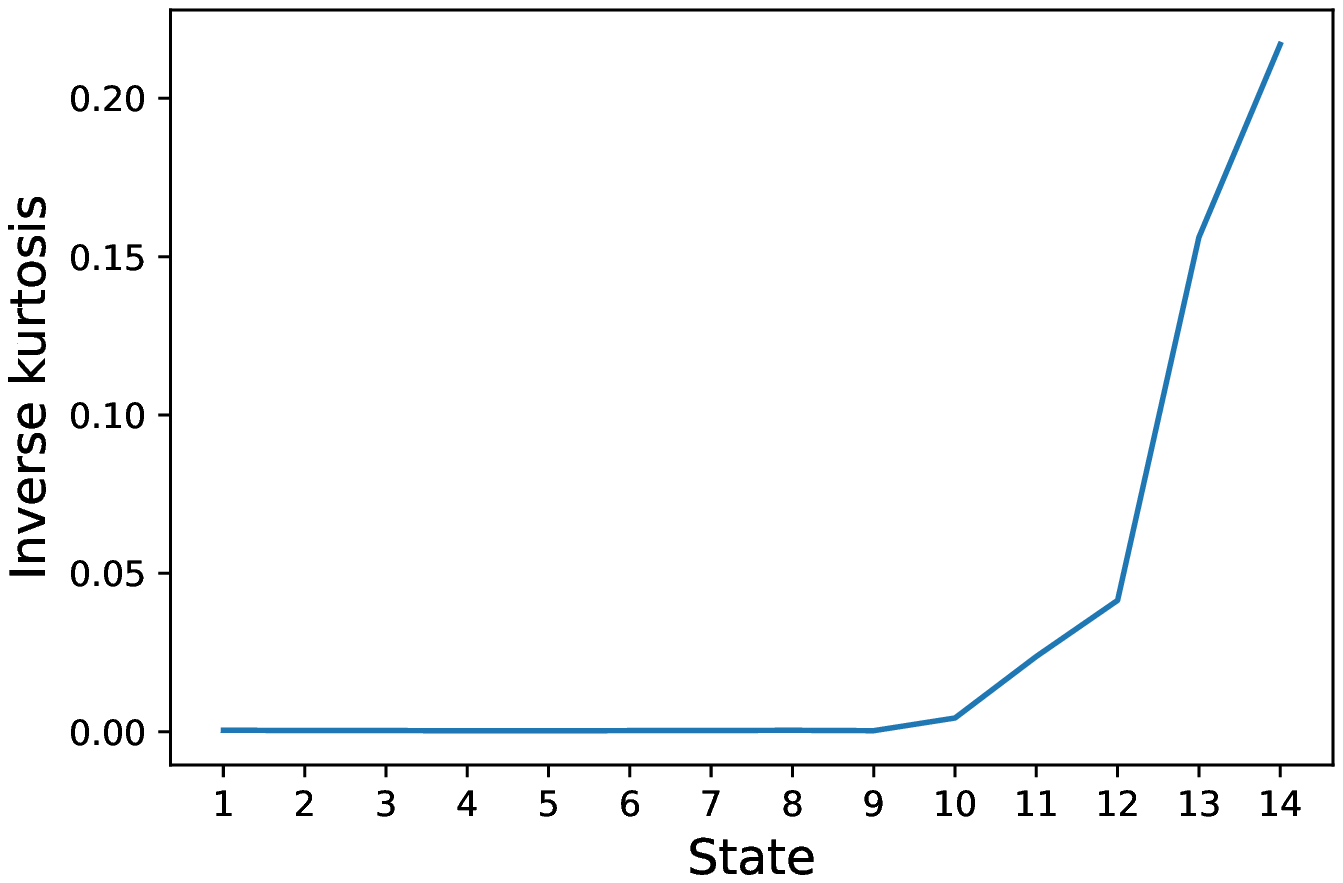}
  \caption{}
  \label{fig:mean_inv_kurts}
\end{subfigure}
\caption{The values of the metric of equation (\ref{eq:metrics}) using inverse kurtosis as $M$, for the accelerations of the three floors of the structure (left) and the average value of all three floors of the metric for the different states (right).}
\label{fig:inv_kurts_mean_inv_kurts}
\end{figure}

The second metric examined is the inverse of the kurtosis of the distributions. Kurtosis is the fourth moment of the distributions and is a metric of how concentrated the mass of the distribution is around its mean. Higher values of kurtosis mean a greater concentration around the mean value and lower values mean more spread values further away from the mean. Therefore, because from its definition, kurtosis is higher for more peaked distributions, its inverse is lower, making it a metric whose value increases as the distributions spread. The results for the inverse kurtosis are presented in a similar way to the standard deviation results in Figure \ref{fig:inv_kurts_mean_inv_kurts}. The results in this case seem even better than in the case of using the standard deviation. For the undamaged states, the metric is almost equal to zero and it monotonically increases only for the second and the third floors. The average also exhibits better results than the standard deviation, being monotonically higher for smaller values of the column-bumper gap. 

\section*{Conclusions}

In the current paper, a metric for detecting and quantifying the nonlinearity of a structure is presented. The metric is based on the calculation of the statistics of the gradients of a model, which is trained as a one-step-ahead model for the data acquired from a structure. The model is built for the baseline undamaged state of a structure, having as input lagged accelerations of the structure and as outputs the values of the acceleration one step ahead in time. For new testing states of the structure the model is recalibrated to the newly-acquired data. The gradients for every output of the model are then calculated with respect to the different inputs of the model for the available data samples of the dataset. The distribution of the values of these gradients is then studied. It is expected that for linear cases the distribution shall be quite peaked and as damage evolves within the structure, it shall cause nonlinear effects and affect the distribution of the gradients. The expected effect is a spread of the values. The spread is studied in the current work using two statistical moments, the standard deviation and the inverse of the kurtosis of the distributions.

\par

The aforementioned methodology is tested on a dataset from a three-storey experimental structure. The structure is tested at 14 different states, nine of which are considered undamaged and the rest five are considered damaged. The difference between the undamaged states is in the stiffness of the columns of the structure and, for the damaged states, the damage is simulated as an added column-bumper setup between two floors of the structure, with varying initial gap between the two elements.

\par

The proposed methodology was applied on the data from the experimental structure and the results revealed that the proposed metrics works as intended. The value of the metric is higher for the damaged cases, making it a tool for identification of damaged states. As the gap between the column and the bumper becomes smaller, making the structure more intensely nonlinear, the metric also increases, which means that the metric could potentially be used for definition of the severity of damage. Furthermore, the metric is higher for the distributions of the gradients of the accelerations of the two floors, between which the column and the bumper are placed, making it also a potential metric for damage localization.

\par

Further validation of the methodology is needed. However, being tested on experimental data the methodology proves quite efficient. Real-life structures which are designed to operate mainly in the linear region of their members exist, e.g.\ nuclear plants. Therefore, a similar situation with the experimental setup of the current work could be encountered in such structures. In these cases a methodology similar to the one presented here could be used for identification, localization and quantification of damage within the structure. Moreover, in future work, analysis of the form of the distributions could be made to infer the type of nonlinearity and other comparisons between the different-state distributions could be made to extract further information about the source of the nonlinearity or even to differentiate the linear cases.

\section*{Acknowledgements}
\label{sec:ack}

G. Tsialiamanis would like to acknowledge the support of the UK EPSRC via the Programme Grant EP/R006768/1 and European Union’s Horizon 2020 research and innovation programme under the Marie Skłodowska-Curie grant agreement No 764547. For the purpose of open access, the authors have applied a Creative Commons Attribution (CC BY) licence to any Author Accepted Manuscript version arising.

% References.
\bibliographystyle{unsrt}
\bibliography{14327_tsi}

\end{document}